\documentclass[journal]{IEEEtran}

\usepackage{ifpdf}
\usepackage{cite}

\ifCLASSINFOpdf
  \usepackage[pdftex]{graphicx}
  \graphicspath{{./pdf/}{./img/}}
  \DeclareGraphicsExtensions{ .pdf, .png, .jpg, .jpeg}
\else
  \usepackage[dvips]{graphicx}
  \graphicspath{{./eps/}}
  \DeclareGraphicsExtensions{.eps}
\fi

\usepackage{amsmath,amssymb}
\usepackage{mathtools}

\usepackage[ruled,vlined,linesnumbered]{algorithm2e}
\usepackage{array}

\ifCLASSOPTIONcompsoc
  \usepackage[caption=false,font=normalsize,labelfont=sf,textfont=sf]{subfig}
\else
  \usepackage[caption=false,font=footnotesize]{subfig}
\fi

\usepackage{fixltx2e}
\usepackage{stfloats}

\ifCLASSOPTIONcaptionsoff
  \usepackage[nomarkers]{endfloat}
 \let\MYoriglatexcaption\caption
 \renewcommand{\caption}[2][\relax]{\MYoriglatexcaption[#2]{#2}}
\fi

\usepackage{url}

\DeclareMathOperator*{\argmax}{arg\,max}

\title{\bf
\LARGE{Bibtex Generator}
}

\usepackage{tabularx}
\usepackage{multirow}
\usepackage{siunitx}
\usepackage{threeparttable}
\usepackage{booktabs,caption}
\usepackage{nomencl}
\usepackage{enumitem}
\usepackage{float}

\usepackage{makecell}

\usepackage{amsthm}

\usepackage[dvipsnames]{xcolor}

\makenomenclature

\begin{document}

% Decalaration page
\begin{table*}[t]
  \centering
  \normalsize
  \begin{tabular}{p{15cm}}
    © 2022 IEEE. Personal use of this material is permitted.  Permission from IEEE must be obtained for all other uses, in any current or future media, including reprinting/republishing this material for advertising or promotional purposes, creating new collective works, for resale or redistribution to servers or lists, or reuse of any copyrighted component of this work in other works. \\
    \\
    Official Publication:\\
    Zuzhao Ye, Yuanqi Gao, and Nanpeng Yu,``Learning to Operate an Electric Vehicle Charging Station Considering Vehicle-grid Integration," IEEE Transactions on Smart Grid, vol. 13, no. 4, pp. 3038-3048, 2022, DOI: 10.1109/TSG.2022.3165479. \\
  \end{tabular}
  \caption*{}
\end{table*}
\setcounter{page}{0}
\newpage

\title{Learning to Operate an Electric Vehicle Charging Station Considering Vehicle-grid Integration}

\author{Zuzhao~Ye,~\IEEEmembership{Graduate Student Member,~IEEE,}
        Yuanqi~Gao,~\IEEEmembership{Member,~IEEE,}
        and~Nanpeng~Yu,~\IEEEmembership{Senior Member,~IEEE}% <-this % stops a space
        
\thanks{Z. Ye, Y. Gao, and N. Yu are with the Department
of Electrical and Computer Engineering, University of Riverside, California,
CA, 92521 USA. e-mail: nyu@ece.ucr.edu.}
}

% make the title area
\maketitle

\begin{abstract}

The rapid adoption of electric vehicles (EVs) calls for the widespread installation of EV charging stations. To maximize the profitability of charging stations, intelligent controllers that provide both charging and electric grid services are in great need. However, it is challenging to determine the optimal charging schedule due to the uncertain arrival time and charging demands of EVs. In this paper, we propose a novel centralized allocation and decentralized execution (CADE) reinforcement learning (RL) framework to maximize the charging station's profit. In the centralized allocation process, EVs are allocated to either the waiting or charging spots. In the decentralized execution process, each charger makes its own charging/discharging decision while learning the action-value functions from a shared replay memory. This CADE framework significantly improves the scalability and sample efficiency of the RL algorithm. Numerical results show that the proposed CADE framework is both computationally efficient and scalable, and significantly outperforms the baseline model predictive control (MPC). We also provide an in-depth analysis of the learned action-value function to explain the inner working of the reinforcement learning agent.
\end{abstract}

\begin{IEEEkeywords}
Electric vehicle, charging station, vehicle-grid integration, reinforcement learning.
\end{IEEEkeywords}

\IEEEpeerreviewmaketitle

\mbox{}

\nomenclature{$T$}{A discrete time horizon.}
\nomenclature{$t$}{$t \in T$, one of the time steps.}
\nomenclature{$\Delta t$}{Length of one time step.}
\nomenclature{$N$}{The total number of parking spots in a charging station.}
\nomenclature{$N^c$/$N^w$}{The total number of chargers/waiting spots.}
%\nomenclature{$N^w$}{The total number of waiting spots.}
\nomenclature{$I$}{The set of EVs that have arrived at the charging station in $T$.}
\nomenclature{$J$}{The set of all chargers in the charging station.}
\nomenclature{$t^a$/$t^d$}{The arrival/departure time of an EV.}
%\nomenclature{$t^d$}{The departure time of an EV.}
\nomenclature{$e^{min}$/$e^{max}$}{The minimum/maximum energy level of an EV.}
\nomenclature{$e^{ini}$/$e^{fnl}$/$e^{tgt}$}{The initial/final/target energy level of an EV.}
%\nomenclature{$e^{tgt}$}{The target energy level specified by the owner of an incoming EV.}
\nomenclature{$Z$}{Profit of the charging station.}
\nomenclature{$B$}{Net revenue of the charging station by charging and discharging EV batteries.}
%\nomenclature{$B^c$}{Net revenue of the charging station by charging EV batteries.}
%\nomenclature{$B^d$}{Net revenue of the charging station by discharging EV batteries.}
\nomenclature{$C^p$}{Penalty paid by the charging station for not charging EVs to their target energy levels.}
\nomenclature{$C^l$}{Demand charge of the charging station.}
\nomenclature{$A$}{The set of all discrete actions of a reinforcement learning agent.}
\nomenclature{$a$}{$a \in A$, one of the actions, i.e. the charging/discharging power selected by a charger for an EV.}
\nomenclature{$a^{min}$/$a^{max}$}{The minimal/maximum charging power. The minimal charging power corresponds to the maximum discharging power.}
%\nomenclature{$a^{max}$}{The maximum charging power.}
\nomenclature{$a^{lower}$/$a^{upper}$}{The lower/upper bound of charging power constrained by battery energy level.}
\nomenclature{$\Delta a$}{Interval between neighboring values in action set $A$.}
\nomenclature{$\beta_{it}$}{A binary variable indicating if the $i$th EV is in the charging station at time $t$.}
\nomenclature{$\eta_{it}$}{A binary variable indicating if the $i$th EV is connected to a charger at time $t$.}
\nomenclature{$m_{t}$}{The net revenue of charging/discharging an EV with 1kWh of electricity.}
\nomenclature{$H$}{The set of time-of-use periods.}
\nomenclature{$h$}{$h \in H$, one of the time-of-use periods.}
\nomenclature{$p^{e}$}{The price of electricity charged by the electric utility.}
\nomenclature{$p^{c}$}{The price paid by an EV customer to the charging station for 1kWh of electricity.}
\nomenclature{$p^{d}$}{The price the charging station pays to EV customers for discharging 1kWh of electricity.}
\nomenclature{$p^{l}$}{The price the charging station pays to the electric utility for 1kW of recorded peak power.}
\nomenclature{$L_{ht}$}{Recorded peak power of a charging station in time-of-use period $h$ at time $t$.}
\nomenclature{$T_B$}{The billing period of demand charge $C_l$.}
\nomenclature{$t^r$}{The remaining dwelling time of an EV.}
\nomenclature{$e^r$}{Remaining energy to be charged for an EV, i.e. the difference between $e$ and $e^{tgt}$.}
\nomenclature{$e_{it}$}{Energy level of the $i$th EV at time $t$.}
\nomenclature{$E^{r,w}$}{Sum of $e^r$ of the EVs in the waiting area.}
\nomenclature{$s$}{The state of environment of a charger.}
\nomenclature{$r$}{The reward received by a charger.}
\nomenclature{$r^b$}{The reward received by a charger through performing charging/discharging of EVs.}
\nomenclature{$r^p$}{The reward received by a charger for not charging an EV to its target energy level.}
\nomenclature{$r^l$}{The reward received by a charger for increasing the recorded peak power.}
\nomenclature{$Q(s,a)$}{The state-action value function of taking action $a$ at state $s$.}
\printnomenclature

\section{Introduction}

\IEEEPARstart{I}{n} the past decade, the global electric vehicle (EV) market has been growing exponentially thanks to the rapid advancement of battery technologies. To support further penetration of EVs, it is critical to develop smart charging stations that could satisfy the charging needs in a cost-effective manner.

The topic of charging station operation optimization has been widely researched with model-based algorithms. The pricing and scheduling of EV charging are jointly considered in the charging station profit maximization problem and solved by an enhanced Lyapunov optimization algorithm \cite{kim2016dynamic}. By jointly considering the admission control, pricing, and charging scheduling, a charging station profit maximization problem is formulated and solved by the combination of multi-sub-process based admission control and gradient-based optimization algorithm \cite{wang2018electrical}. The planning and operation problem of single output multiple cables charging spots is formulated and solved by a two-stage stochastic programming \cite{zhang2016optimal}. The charging station scheduling problem that considers vehicle-to-vehicle charging is formulated as a constrained mixed-integer linear program and solved by dual and Benders decomposition \cite{you2014efficient}. The scheduling of battery replacement in charging stations is solved by a multistage stochastic programming algorithm \cite{dong2015phev}. By considering renewable generation and energy storage systems in the design \cite{huang2015matching, dominguez2019design,deng2020operational} and operation of charging stations \cite{yan2018optimized,awad2016optimal}, significant economic benefits can be gained.

Model-based control approaches have also been applied to schedule EV charging considering vehicle-grid integration (VGI). A decentralized control algorithm to optimally schedule EV charging is developed to fill the valleys in electric load profiles \cite{gan2012optimal}. To mitigate the impact on distribution networks, the EV charging and discharging coordination problem is solved with the strategy of the coalition formation \cite{yu2014phev}. To smooth out the power fluctuations caused by distributed renewable energy sources, a joint iterative optimization method is developed to schedule the charging of EVs and minimize the cost of load regulation \cite{wei2020aggregation}. The problem of routing and charging scheduling of EVs considering VGI is formulated as an optimization problem solved by an approximated algorithm \cite{tang2018distributed}. The VGI, vehicle-to-vehicle energy transfer, and demand-side management in vehicle-to-building aspects have also been studied with the mixed integer programming \cite{koufakis2016towards} and a distributed control algorithm \cite{nguyen2012optimal}.

The aforementioned model-based methods rely heavily on sophisticated algorithm designs that are tailored for specific charging scenarios. Most of the model-based methods cast the charging station scheduling problem as a complex optimization problem, which is time-consuming to solve when the scale of the charging station increases. To address the scalability issues of model-based methods, researchers are turning to reinforcement learning (RL), which excels at solving sequential decision-making problems in real-time. The RL agent aims at learning a good control policy by interacting with the environment or a simulation system to achieve the best long-term rewards. Combining RL algorithms with deep neural networks has led to breakthroughs in solving complex problems with large-scale state and action spaces \cite{dulac2015deep}.

\textcolor{black}{The application of RL to schedule the charging of a single EV is widely covered in the literature.} An early attempt leverages the tabular Q-learning method \cite{dimitrov2014reinforcement}, which can only deal with small-scale state and action space. To overcome the limitations of tabular methods, kernel averaging regression functions are used to approximate the action-value functions in scheduling the charging of an individual EV \cite{chics2016reinforcement}. \cite{wan2018model} further extends the charging scheduling problem for a single EV by considering bidirectional energy transfer and utilizing deep neural networks to approximate state-action values. \textcolor{black}{Apart from reducing the charging cost, drivers’ anxiety about the EVs' range \cite{yan2021deep} and photovoltaic self-consumption \cite{dorokhova2021deep} can also be quantified and included as training targets. To ensure the EV is sufficiently charged upon departure, \cite{chang2019charging} introduces an exponential penalty term to encourage the charger finishing the charging task on time, while \cite{li2019constrained} utilizes constrained policy optimization to learn a safe policy without the need to manually design a penalty term. Long short-term memory (LSTM) is also widely used to predict the electricity price as an input to the RL algorithm \cite{wan2018model,zhang2020cddpg, wang2020autonomous, li2021electric}.} 

\textcolor{black}{Compared to the single EV algorithms,} it is much more challenging to apply RL to coordinate the charging of multiple EVs in a charging station. This is because the dimensionality of the state space is changing with the stochastic arrival and departure of EVs. To address this problem, a two-stage heuristic RL algorithm is proposed \cite{vandael2015reinforcement}. In the first stage, an RL agent learns the collective EV fleet charging strategy. In the second stage, a simple heuristic is used to translate collective control action back to individual charger control action. The drawback of this approach is that the simple heuristic cannot guarantee optimal charging scheduling. An alternative approach to tackle the difficulty of dimension-varying state and action space is to represent the state-action function using a linear combination of a set of carefully designed feature functions \cite{wang2019reinforcement}. \textcolor{black}{However, designing good feature functions is a manual process that can be challenging. To overcome this challenge, \cite{sadeghianpourhamami2019definition} develops a novel grid representation of EVs' required and remaining charging time such that the dimensions of state and action are constants. The problem with this grid approach is that the charger can only select to charge with full power or to not charge, without any intermediate power outputs. While in most of the research papers RL agents are deployed to determine output charging powers, \cite{ding2020optimal} instead uses RL to predict future conditions. A model-based method is used determine charging powers. The drawback of this approach is that the prediction can hardly be accurate. Researchers also studied the problem of charging scheduling of multiple individual EVs in a local area to reduce the load of the grid \cite{da2019coordination,tuchnitz2021development}. In these work, each charger is an RL agent and collective features (e.g. the total load) are usually introduced to the state vector.} These pioneering studies demonstrate the potential of RL in charging scheduling and pave the way for further research. 

\textcolor{black}{Other interesting applications of RL related to charging include the scheduling of multiple charging stations where each charging station is treated as a unit \cite{shin2019cooperative,lee2021dynamic}, the dynamic pricing for charging services \cite{zhao2021dynamic, abdalrahman2020dynamic, moghaddam2020online}, and the scheduling of battery swapping stations \cite{gao2020deep, wang2020vehicle}.}

In this paper, we propose an innovative centralized allocation and decentralized execution (CADE) framework to schedule the charging activities at the charging station. \textcolor{black}{Compared to previous work \cite{vandael2015reinforcement, wang2019reinforcement, sadeghianpourhamami2019definition, ding2020optimal}, the problem formulation in this paper considers waiting area, VGI, and demand charge, making it more realistic and comprehensive.} By synergistically combining the merits of RL and advanced optimization, our proposed CADE framework is highly scalable and yields competitive charging station profit compared to state-of-the-art baseline algorithms. The centralized allocation module leverages an optimization technique to determine which EVs should be connected to the chargers. To address the challenge of the dimension-varying state and action space of the charging station, we propose a decentralized execution scheme, where each charger is represented by an RL agent and determines its own charging/discharging actions. The individual chargers will submit operational experiences to a shared replay memory, which provides training samples to update the policies of the RL agents. Most of the existing algorithms neglect the demand charge cost as it is extremely difficult to predict the peak demand of a charging station ahead of time. In our proposed framework, an online algorithm is designed to assign the charging station-wide penalty associated with demand charge to the individual chargers.

The contributions of this paper are summarized below:
\begin{enumerate}

\item We propose a scalable centralized allocation and decentralized execution (CADE) framework to operate an EV charging station considering waiting area, vehicle-grid integration, and demand charge. By synergistically combining advanced optimization with reinforcement learning algorithms, our proposed approach identifies both the EV assignment and the chargers' outputs that yield high charging station profit.

\item By modeling each charger as an intelligent RL agent, our proposed algorithm tackles the challenge of dealing with dimension-varying state and action space associated with stochastic EV arrivals and charging needs. To improve the learning efficiency, we allow the chargers to exploit a shared replay memory of operational experience in learning decentralized charger control policies.

\item Comprehensive numerical studies demonstrate that our proposed CADE framework is not only extremely scalable but also outperforms state-of-the-art baseline algorithms in terms of charging station profit.

\end{enumerate}

The rest of this paper is structured as follows. Section \ref{sec:problem formulation} formulates the charging scheduling problem as an optimization problem. Then, Section \ref{sec:tech-method} reformulates the charging scheduling problem as a Markov Decision Process and proposes the CADE framework to solve the sequential decision making problem. Case studies are included in Section \ref{sec:case} to demonstrate the effectiveness of our proposed method. Section \ref{sec:conclusion} provides a brief conclusion.

\section{Problem Formulation}
\label{sec:problem formulation}

We will first formulate the charging station scheduling problem as an optimization problem. Let us consider a charging station with $N$ parking spots, among which $N^c$ are charging spots and $N^w$ are waiting spots, such that $N=N^w+N^c$. When the number of EVs in the charging station is larger than the number of chargers, the charging station needs to determine which EVs shall be connected to the chargers. Here, we assume that the time of switching an EV from a waiting spot to a charger or vice versa is negligible.

The goal of the charging station's scheduling system is to maximize its profit while satisfying EV customers' charging needs. The operating profit of a charging station $Z$ consists of three components: the net revenue of charging and discharging EVs, $B$, the penalty of failing to satisfy the customers' charging needs, $C^{p}$, and the demand charge of the station, $C^{l}$ as shown in \eqref{eq:profit}.
\begin{equation}
    Z = B - C^{p} - C^{l} 
\label{eq:profit}
\end{equation}

We assume that $Z$ is measured over a time horizon $T$, which consists of intervals of equal length $\Delta t$. All of the incoming EVs during $T$ are denoted by a set $I$. We will describe each component of $Z$ in detail in the next few subsections.

\subsection{Net Revenue from Charging and Discharging EVs}
The EV owners pay the charging station $p_t^{c}$ for each kWh of electricity charged and receive $p_t^{d}$ for each kWh of electricity discharged from the batteries, which was sold to the electric grid. The charging station pays or receives $p_t^{e}$ for each kWh of energy bought from or sold to the electric utility.

Suppose at time $t$, the $i$th EV is charged/discharged at a power level $a_{it}$, where $a_{it} > 0$ means charging and $a_{it} < 0$ means discharging. The net revenue from charging and discharging EVs, $B$, can be calculated as:

\begin{equation}
  B = \sum_{i \in I} \sum_{t \in T}m_t|a_{it}|\Delta t,
\label{eq:B}
\end{equation} where $m_{t}$ is the net revenue of charging/discharging an EV with 1kWh of energy at time $t$ and can be calculated by:
\begin{equation}
m_t =
\begin{cases}
    p^{c}_{t} - p^{e}_{t} & \text{if $a_{it} \geq 0$}\\
    p^{e}_{t} - p^{d}_{t} & \text{if $a_{it} < 0$}
\end{cases}
\label{eq:mt}
\end{equation}

Note that the price paid to the customer for discharging the EV battery is higher than the EV charging cost, i.e. $p^{d}_{t} > p^{c}_{t}$, $\forall t$. This is necessary to cover the degradation cost of the EV battery \cite{foggo2017improved}.

\subsection{Penalty of Failing to Satisfy Customers' Charging Needs}

Upon arrival, an EV customer $i$ will specify the departure time $t^{d}_i$ and the desired target energy level $e^{tgt}_i$ at departure. The charging station will try to ensure that the EV's energy level is not lower than $e^{tgt}_i$ at time $t^d_i$. If the target energy level is not met, a penalty will be imposed on the charging station to compensate the customer. The penalty is denoted as $c^p_i$.

 Such a penalty shall reflect the gap between the final energy level $e^{fnl}_i$ and the target energy level $e^{tgt}_i$. For simplicity, we use a linear function to calculate $c^p_i$ as suggested by \cite{tuchnitz2021development}:
\begin{equation}
   c^{p}_{i} = \mu(e^{tgt}_{i} - e^{fnl}_{i})^+,
\label{eq:cpi}
\end{equation} where $\mu$ is a constant coefficient and $x^+=max(x,0)$. Note that a penalty is triggered only when $e^{fnl}_{i} < e^{tgt}_{i}$. $e^{fnl}_i$ can be derived by \eqref{eq:efi}:
\begin{equation}
   e^{fnl}_i = e^{ini}_i + \sum_{t = t^a_i}^{t^d_i}a_{it}\Delta T,
\label{eq:efi}
\end{equation} where $e^{ini}_i$ is the initial energy level of the $i$th EV and $t^{a}_i$ is its arrival time. Also note that an EV customer cannot specify an $e^{tgt}$ higher than a threshold that is physically achievable with the charging infrastructure. We explicitly express this requirement as shown in \eqref{eq:etgt}:

\begin{equation}
   e^{tgt}_i \le min\left [e^{max}, e^{ini}_i + a^{max}(t^d_i-t^a_i) \right ],
\label{eq:etgt}
\end{equation}
where $e^{max}$ is the maximum capacity of the battery, and $a^{max}$ is the maximal charging power.

The total penalty for a charging station $C^p$ is the sum of $c^p_i$ over all EVs as shown in \eqref{eq:Cp}:

\begin{equation}
C^{p} = \sum_{i \in I} c^{p}_{i}
\label{eq:Cp}
\end{equation}
%\subsection{Demand Charge $C^{l}$}
\subsection{Demand Charge}
Another important component of the charging station operation cost is the demand charge, which is calculated based on the maximum load in the billing period $T^B$. Note that the demand charge during on-peak and off-peak periods can be drastically different, which incentivizes charging stations to charge EV batteries during off-peak hours and discharge EV batteries during on-peak hours.

Let us denote the set of time-of-use periods to be $H$, and $h \in H$ is one of the time-of-use periods. The price to be paid for each kW of peak demand in time-of-use period $h$ is $p^{l}_{h}$. The charging station's demand charge $C^{l}$ can be calculated by \eqref{eq:Cl}: 
\begin{equation}
C^{l} = \frac{T}{T^B}\sum_{h \in H} p^{l}_{h} \cdot L_{h}^+,
\label{eq:Cl}
\end{equation} where $T/T_B$ serves as a scaling factor if we choose a simulation time horizon $T$ that is longer or shorter than $T_B$. $L_{h}$ is the peak load recorded in time-of-use period $h$ during $T$. The electric utility does not compensate the customer when the peak demand is negative. $L_h$ is calculated by summing up the charging/discharging power of all the chargers at a given time $t \in T_h$ and taking the maximal value among the results, as shown in \eqref{eq:Lh}:
\begin{equation}
L_{h} = \max_{t \in T_{h}} \left ( \sum_{i \in I} a_{it}\right ),
\label{eq:Lh}
\end{equation}
where $T_{h} = f(h)$ and $f: H\to T$, i.e. $T_{h}$ is the set of time steps in time-of-use period $h$.

\subsection{Summary of the Charging Station Optimization Problem}
\label{subsec:op}
The charging station scheduling problem can be formulated as the following optimization problem, which aims at maximizing the profit $Z$ \eqref{eq:op} while satisfying the operational constraints \eqref{eq:cons-1} - \eqref{eq:cons-7}:

\begin{equation}
\begin{split}
    \underset{a_{it},\eta_{it}}{\text{max}} \ Z = & \sum_{i \in I}\sum_{t \in T}m_t|a_{it}|\Delta t \\
    & - \sum_{i \in I} \mu \left [e^{tgt}_{i} - \left (e^{ini}_i + \sum_{t = t^a_i}^{t^d_i}a_{it}\Delta T \right ) \right ] ^+ \\
    & - \frac{T}{T_B}\sum_{h \in H} p^{l}_{h} \cdot \left [ \max_{t \in T_{h}} \left ( \sum_{i \in I} a_{it}\right ) \right ] ^+,
\end{split}
\label{eq:op}
\end{equation}

subject to:
\begin{subequations}
\begin{alignat}{2}
a^{min} \leq a_{it} \leq a^{max}, &\quad\forall i \in I, \forall t \in T \label{eq:cons-1}\\
a_{it} = 0, \ \text{if $\eta_{it}$ = 0}, &\quad\forall i \in I, \forall t \in T \label{eq:cons-2}\\
\eta_{it} \in \{0, 1 \}, &\quad\forall i \in I, \forall t \in T \label{eq:cons-3}\\
\sum_{i \in I}\eta_{it} \leq N^c, &\quad\forall t \in T \label{eq:cons-4}\\
\sum_{i \in I}\beta_{it} - \sum_{i \in I}\eta_{it} \leq N^w, &\quad\forall t \in T \label{eq:cons-5}\\
\beta_{it} - \eta_{it} \geq 0, &\quad\forall i \in I, \forall t \in T \label{eq:cons-6}\\
e^{min} \leq e^{ini}_{i} + \sum_{t=t^{a}_{i}}^{t'} a_{it}\Delta t  \leq e^{max}, &\forall i \in I, \forall t' \in \left[t^{a}_{i}, t^{d}_{i}\right]
\label{eq:cons-7}
\end{alignat}
\end{subequations}

The operational constraints can be grouped into the following three categories.

\subsubsection{Power constraints}
The feasible range of charging power for an EV is defined by \eqref{eq:cons-1}. Equation \eqref{eq:cons-2} states that an EV shall receive zero power if it is not connected to a charger, where $\eta_{it}$ is a binary variable indicating if the $i$th EV is connected to a charger at time $t$. $\eta_{it} = 1$ indicates that the EV is connected to a charger.

\subsubsection{Space constraints}
The total number of EVs connected to chargers at the same time is limited by \eqref{eq:cons-4}. Equation \eqref{eq:cons-5} limits the number of EVs in the waiting spots, where $\beta_{it}$ is a binary variable indicating if the $i$th EV is in the charging station at time $t$. $\beta_{it}=1$ means that the EV is in the charging station. If an EV is in the charging station, it is either connected to a charger or in a waiting spot. Equation \eqref{eq:cons-6} implies that an EV cannot be connected to a charger if it is not in the charging station.

\subsubsection{Energy constraints}
The battery energy level of an EV must fall within the feasible range of $[e^{min}, e^{max}]$ when it is in the charging station as specified in \eqref{eq:cons-7}.

If all future information of EVs, i.e. $\beta_{it}$, $e^{ini}_{i}$, $e^{tgt}_{i}$, $t^{a}_{i}$, and $t^{d}_{i}$ $\forall i \in I, \forall t \in T$ are known, then an oracle is able to obtain the global optimal solution of the charging station scheduling problem. In this case, the charging station scheduling problem is formulated as a mixed integer programming (MIP) problem.

In practice, it is impossible to have perfect predictions of arriving EVs. Instead, model predictive control methods are usually developed to handle similar problems \cite{kou2015stochastic,tang2016model}. However, the effectiveness of the MPC-based algorithm heavily relies on accurate predictions of future EV arrivals and does not scale well with the size of the problem. This motivates us to develop model-free methods, such as reinforcement learning-based algorithms, which is capable of making real-time decisions in complex environments.

\section{Technical Methods}
\label{sec:tech-method}

\begin{figure*}[!t]
\centering
\includegraphics[width=0.95\textwidth]{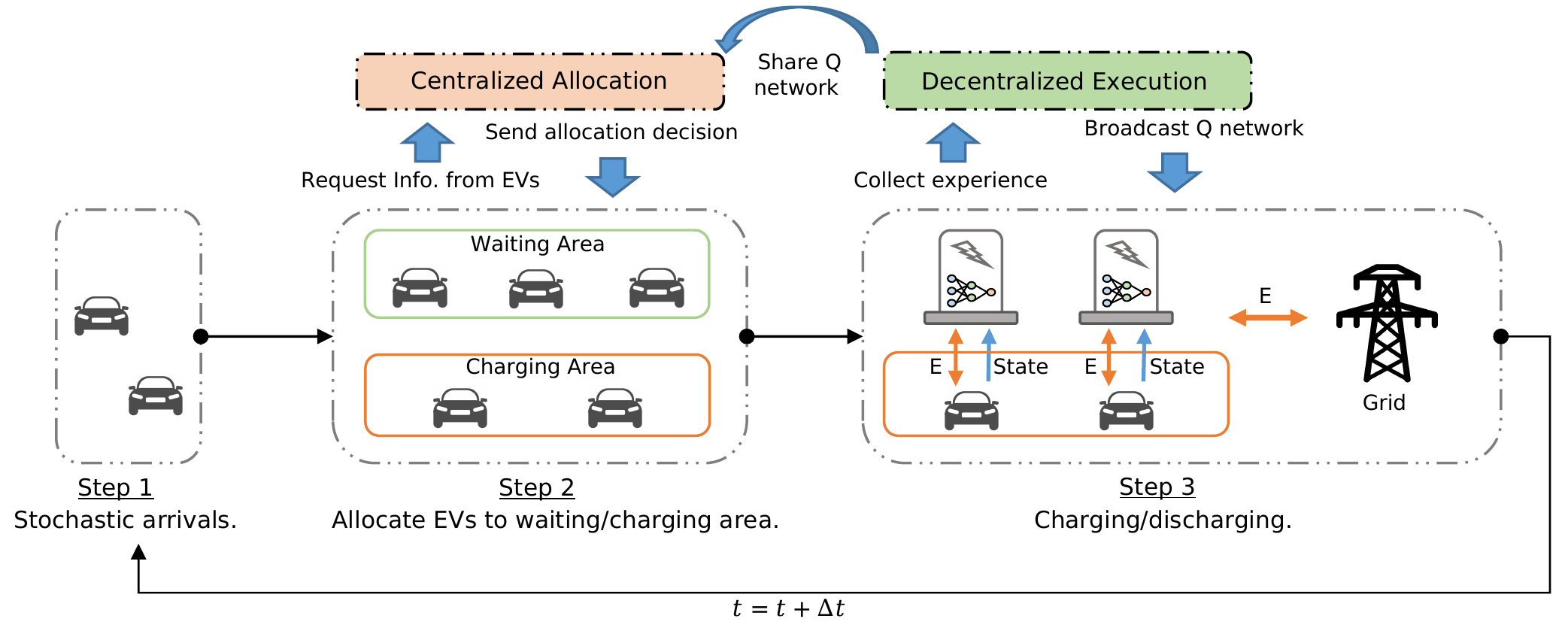}
\caption{Overview of the centralized allocation and decentralized execution (CADE) framework.}
\label{fig:overview}
\end{figure*}
We propose a centralized allocation and decentralized execution framework (CADE) to solve the charging station scheduling problem formulated in Section \ref{sec:problem formulation}. The overall framework of CADE for charging station scheduling is shown in Fig. \ref{fig:overview}. EVs stochastically arrive at the charging station. Immediately after observing the actual EV arrival at time step $t$, a centralized allocation algorithm determines which EVs shall be connected to chargers or assigned to the waiting area. Then each charger acts as an individual agent and makes charging/discharging decisions based on the state of the EV connected with it. The sequential decision-making problem of each charger is formulated as a Markov Decision Process (MDP). The operational experiences of all chargers will be collected and used in a learning process. In this process, the parameters of the state-action value function's neural networks will be updated and then broadcast to all chargers. 

\subsection{Overview of Markov Decision Process}
In this subsection, we briefly introduce MDP, which is the mathematical foundation for sequential decision-making problems. In an MDP, an agent interacts with an environment over a sequence of discrete time steps $t=\{1,2,...,\mathcal{T}\}$. At each time step $t$, the agent senses the state $s \in S$ of the environment and takes action $a \in A$ accordingly. Once the action is taken, the environment will transition to a new state $s^{\prime}$ with probability $Pr(s^{\prime}|s,a)$ and provide a reward $r$ to the agent. The agent's goal is to maximize the expected sum of discounted rewards $R_t = \mathbb{E}\left[\sum_{\tau=t}^{\mathcal{T}}\gamma^{\tau - t} r_\tau\right]$, where $\gamma \in [0, 1]$ is a discount factor that balances the relative importance of near-term and future rewards. 

Suppose an agent follows a policy $\pi$ to make decisions, we can define the state-action value function of the policy in \eqref{eq:Qpi} to quantify how good it is to take an action $a$ in a state $s$:

\begin{equation}
Q^{\pi}(s,a) = \mathop{\mathbb{E}} \left [ R_t | s_t = s, a_t = a, \pi\right ]
\label{eq:Qpi}
\end{equation}

The state-action value function must satisfy the following Bellman equations \eqref{eq:Qpid}:

\begin{equation}
   Q^{\pi}(s,a)= \mathop{\mathbb{E}_{s^{\prime}}} \left [ r + \gamma \mathop{\mathbb{E}_{a^{\prime} \sim \pi(s^{\prime})}} Q^{\pi}(s^{\prime}, a^{\prime}) | s, a, \pi\right ]
\label{eq:Qpid}
\end{equation}

We define the optimal state-action value function $Q^*(s,a)$ as $Q^*(s,a) = \max_{\pi}Q^{\pi}(s,a)$. Then the optimal deterministic policy $\pi^*$ is the one that chooses actions $a^* = \text{argmax}_{a \in A}Q^*(s, a)$.

\subsection{Formulate the Charger Control Problem as an MDP }
In the individual charger control problem, each charger is treated as an agent who interacts with the environment, which consists of the EV connected to the charger and the charging station.

Suppose the set of chargers is $J$. For an individual charger $j \in J$, its state of environment at time $t$ is defined as $s_{jt} = \{\delta_{jt}, \ t, \ t^{r}_j, \ e_{jt}, \ e^{r}_{jt}, \ N^{EV,w}_t, \ E^{r,w}_t, \ \Tilde h_t, \ L_{ht} \}$. $\delta_{jt} \in \{0, 1\}$ indicates if there is an EV connected to charger $j$. $t \in T$ denotes the global time. $t^{r}_j = t^{d}_j - t$ is the remaining dwelling time of the EV connected to the charger. $e_{jt}$ denotes the current energy level of the EV battery. $e^{r}_{jt} = e^{tgt}_j - e_{jt}$ is the remaining energy to be charged for the EV battery. $N^{EV,w}_t$ is the total number of EVs in the waiting area at time $t$. $E^{r,w}_{t}$ is the sum of remaining energy to be charged for all EVs in the waiting area, which can be calculated as $E^{r,w}_{t} = \sum_{\substack{k = 1 }}^{N^{EV,w}_t} e^{r}_{kt}$. $\Tilde h_t$ is an one-hot encoded vector indicating the current time-of-use period. The dimension of $\Tilde h_t$ is $dim(H)$. For example, if there are three time-of-use periods and we are currently in the second period, then $\Tilde h_t = [0, 1, 0]$. $L_{ht}$ is the maximum load that the charging station has seen so far for time-of-use period $h$, which corresponds to global time $t$.

Note that $t^{r}_j$ and $e^{r}_{jt}$ quantify the urgency of the need to charge the EV connected to charger $j$. $N^{EV,w}_t$ and $E^{r,w}_{t}$ quantify the urgency of the need to charge the EVs in the waiting area. The agent can obtain the current time-of-use pricing for electricity given the global time $t$.

The action for a charger $j$ at time $t$ is its output power $a_{jt}$. In this paper, we discretize the action space using discrete charging rates as suggested by \cite{binetti2015scalable, sun2016optimal}. The upper bound of the charger power output action $a^{upper}_{jt}$ can be calculated as the minimum of the physical maximum charger output and the maximum charging rate allowed to fill up the battery of the EV: $a^{upper}_{jt} = \min (a^{max}, \frac{e^{max}-e_{jt}}{\Delta t})$. The lower bound of the charger power output action $a^{lower}_{jt}$ can be calculated as the maximum of the physical minimum charger output and the maximum discharging rate allowed to deplete the battery of the EV: $a^{lower}_{jt} = \max (a^{min}, \frac{e^{min}-e_{jt}}{\Delta t})$. Finally, the feasible action space for a charge is defined as an ordered set of discrete and increasing charger output $A_{jt} = \{ a^{lower}_{jt}, ..., a^{upper}_{jt} \}$ with uniform difference $\Delta a$ between adjacent actions.

%******************
% 1. Rb
The design of the reward function for the chargers shall be consistent with the charging station operation objective in \eqref{eq:profit}, which is maximizing the net revenue $B$ from charging/discharging EVs and minimizing charging station demand charge $C^l$ and the penalty of not satisfying customers' charging needs $C^p$. The reward function of a charger equals to the sum of three components: $r_{jt} = r^{b}_{jt} + r^{p}_{jt} + r^{l}_{jt}$. The first component $r^{b}_{jt} = m_{t}|a_{jt}|\Delta t$ represents the net revenue of a charger for charging/discharging an EV in the period of $\Delta t$.

The second component $r^p_{jt}=-(c^{p}_{jt} + c^{p,w}_{jt})$ accounts for the penalty of failing to satisfy the customer's charging needs. The second component of the reward function has two parts. The first part $c^{p}_{jt}$ is a penalty assigned to the charger $j$ when the EV connected to it does not reach its desired energy level upon leaving, which can be calculated as follows:
\begin{equation}
c^{p}_{jt} =
\begin{cases}
    \mu(e^{tgt}_j-e_{jt})^+ & \text{if $t^r = 0$}\\
    0 & \text{if $t^r > 0$}
\end{cases}
\label{eq:cp}
\end{equation}
The second part $c^{p,w}_{jt}$ is a penalty assigned to the charger $j$ when EVs in the waiting area do not reach their desired energy level upon leaving. Since EVs in the waiting area are not directly associated with a specific charger, the corresponding penalty will be split among all chargers in the charging station. The aggregated penalty $R^{p,w}_t$ for all EVs in the waiting area can be calculated as $R^{p,w}_t = \sum_{k = 1}^{N^{EV,w}_t} c^p_{kt}$. To encourage chargers to increase charging power when the EVs in the waiting areas have not reached their desired energy level, we propose splitting $R^{p,w}_t$ in a way that assigns chargers fewer penalties if their power output levels are higher:

\begin{equation}
    c^{p,w}_{jt} = \frac{a^{max}-a_{jt}}{\sum_{j \in J}(a^{max}-a_{jt})}R^{p,w}_t
\label{eq:cp-split}
\end{equation}

The third component of the reward function $r^l_{jt}$ is associated with the demand charge of the charging station. The aggregated additional demand charge to be paid by the charging station $R^l_t$ at time $t$ is positive when the total charging power $\sum_{j \in J} a_{jt}$ exceeds the maximum load of the charging station $L_{ht}$ up to time $t$. $R^l_t$ can be calculated by following Algorithm \ref{algo:Rl}.
\begin{algorithm}[h!]
\SetAlgoLined
Obtain maximum charging station load $L_{ht}$ for time-of-use period $h$ up to time $t$\;
Initialize $R^l_t = 0$\;
\If{$\sum_{j \in J} a_{jt} > L_{ht}$}
    {
    \BlankLine
    $R^{l}_t \leftarrow - \frac{T}{T_B} p^{l}_{h} \left(\sum_{j \in J} a_{jt} - L_{ht} \right)$\;
    \BlankLine
    $L_{ht} \leftarrow \sum_{j \in J} a_{jt}$\;
   }
\caption{Online calculation procedure for $R^{l}_t$}
\label{algo:Rl}
\end{algorithm}

If a charger $j$ selects a higher charging power $a_{jt}$, then it contributes more to the increase in demand charge. Thus, we design $r^l_{jt}$ in a way that assigns a charger with higher power output a greater penalty as shown in \eqref{eq:rl}:
\begin{equation}
    r^l_{jt} =  \frac{a_{jt}}{\sum_{j \in J} a_{jt}} R^l_t,
\label{eq:rl}
\end{equation}

Note that if the study time horizon $T$ is different from the electricity billing period $T_B$, then the increase in demand charge is scaled accordingly in Algorithm \ref{algo:Rl}.

\subsection{Decentralized Execution at the Charger Level}
The chargers in the charging station determine the charging/discharging power for each time step by following an RL-based decentralized execution framework. Individual chargers collect historical operational experiences, which include the current state, the action, the next state, and the reward, and store them in a shared replay memory $M$. The action value functions corresponding to the chargers will be learned through a deep Q-learning (DQN) algorithm \cite{mnih2015human} using the shared replay memory. DQN is selected due to its ability to handle large state and action spaces. The learned parameters of the Q networks will be periodically broadcasted to the individual chargers. In real-time operations, the chargers will be taking charging/discharging actions based on its own observation of the state in a decentralized manner.

In the DQN algorithm, a deep neural network parameterized by $\theta$ is used to approximate the action value function $Q$. To stabilize the learning process, a second neural network parameterized by $\theta^{-}$ called the target network is introduced to reduce the strong correlations between the function approximator and its target. The parameters of the Q-network are iteratively updated by stochastic gradient descent on the following loss function: 

\begin{equation}
L(\theta) = \mathop{\mathbb{E}} \left [r + \gamma (1- d) \max_{a^{\prime} \in A} Q(s^{\prime}, a^{\prime}; \theta^-) - Q(s,a;\theta) \right ]^2,
\label{eq:loss}
\end{equation}
where $s$ and $s^{\prime}$ are the current state and the next state of the environment, respectively. $d \in \{0,1\}$ is an indicator of the terminal state for each episode, and $d = 1$ indicates $s^{\prime}$ is the terminal state. The target net's parameters $\theta^-$ are replaced by $\theta$ after every certain number of training episodes.

We design the Q-network with the following structure. The input layer of the neural network includes the state and action variables $(s,a)$ of the individual chargers. The output layer has a single neuron representing the value of $Q(s,a)$. This design is suitable for the charger decision-making problem as the charging powers are ordinal variables.

\subsection{Centralized Allocation at the Charging Station Level}
At each time step, the charging station scheduler needs to allocate the EVs to the chargers and waiting spots to maximize its own profit. Instead of directly solving the optimization problem formulated in Section \ref{subsec:op}, we propose a centralized EV allocation algorithm that leverages the learned action value function $Q$ for the chargers \textcolor{black}{in the previous time step.}

Suppose at time $t$, we have $N^{EV}_t$ EVs in the charging station. An EV can either be connected to one of the chargers or parked in the waiting area. Let a binary variable $\alpha_k$ denotes whether the $k$th EV is connected to a charger. $\alpha_k=1$ indicates that the $k$th EV is connected to a charger, and such a connection will create an action value $q^{c}_k = max_{a}Q(s(\alpha_k=1), a)$, where \textcolor{black}{$Q$ is the action value function learned in the previous time step} and $s(\alpha_k=1)$ denotes the state of the charger when it is connected to the $k$th EV. On the other hand, when the $k$th EV is parked on a waiting spot, it is equivalent to be connected to a charger with zero power output, in which case, the connection will create an action value $q^{w}_k = Q(s(\alpha_k=1), a=0)$. As a summary, the action value created by the connection of the $k$th EV is $\alpha_k q^{c}_k + (1-\alpha_k)q^{w}_k$. Note that for state vector $s(\alpha_k=1), \forall k \in \{1, 2, ..., N^{EV}_t\}$, the remaining energy to be charged for all EVs in the waiting area can be approximated by its value in the previous time step. 

The charging station's scheduling problem can be solved by finding the EV allocation that maximize the summation of the action values created by the connections of all EVs while satisfying the space limitations as shown in \eqref{eq:ca}, \eqref{eq:ca-cons-1}, and \eqref{eq:ca-cons-2}:

\begin{equation}
\begin{split}
    \max_{\alpha_k}  \sum_{k = 1}^{N^{EV}_t} \alpha_k q^{c}_k + (1-\alpha_k)q^{w}_k
\label{eq:ca}
\end{split}
\end{equation}

subject to:
\begin{subequations}
\begin{alignat}{2}
& \sum_{k = 1}^{N^{EV}_t}\alpha_k = \min \left ( N^c, N^{EV}_t \right ), \  \label{eq:ca-cons-1}\\
& \alpha_k \in \{ 0, 1 \}, \quad \forall k \in \{1, 2, ..., N^{EV}_t\}, \label{eq:ca-cons-2}
\end{alignat}
\end{subequations}

Equation \eqref{eq:ca-cons-1} enforces the constraint that at most $N^c$ EVs are connected to chargers if $N^{EV}_t > N^c$. If $N^{EV}_t \leq N^c$, then all EVs should be allocated to charging area.

\begin{algorithm}[h!]
\SetAlgoLined
Initialize exploration rate $\epsilon = 1$\;
Initialize replay memory M\;
Initialize state-action value function network $Q(s, a; \theta)$ with random weights $\theta$\;
Initialize target network $Q(s, a, \theta^{-})$ with $\theta^- = \theta$\;

\For{episode = 1:n}{
Initialize $t = 0$ and $N^{EV}_t = 0$\;
Initialize maximum loads $L_{ht} = 0$, $\forall h \in H$\;

\While{{$t < T$}}{
    If $N^{EV}_t < N$, admit arriving EVs to the charging station\;
    Allocate EVs to chargers \& waiting area by solving optimization problem \eqref{eq:ca} - \eqref{eq:ca-cons-2} \textcolor{black}{where $Q(s, a; \theta)$ is utilized to generate $q^{c}_k$ and $q^{w}_k, \forall k \in \{1, 2, ..., N^{EV}_t\}$}\;
    \For{$j \in J$}{
    	Obtain $s_{jt}$ for charger $j$\;
        Draw $x$ from uniform distribution $U(0, 1)$\;
        \eIf{$x < \epsilon$}
        {
            Randomly select $a_{jt} \in A_{jt}$
        }
        {
            Select $a^*_{jt} = \argmax_{a_{jt}} Q(s_{jt}, a_{jt}; \theta) $\;
        }
        Calculate $r^b_{jt}$\;
        The environment transitions to the next state $s_{j(t+1)}$\;
    }
    
    \For{$j \in J$}{
    	Calculate $r^p_{jt}$, $r^l_{jt}$, and finally $r_{jt}$\;
        Add $(s_{jt}, a_{jt}, s_{j(t+1)}, r_{jt})$ to M\;
    }
    Sample a mini-batch $m$ from M\;
    Update $\theta$ by using the gradient descent method to minimize the loss function in \eqref{eq:loss}\;
    Decrease exploration rate $\epsilon$\;
    Set $\theta^- = \theta$ every $K$ episodes\;
    $t \mathrel{{+}{=}} \Delta t$\;
}
}
\caption{Summary of the CADE Framework}
\label{algo:CADL}
\end{algorithm}
\subsection{Summary of CADE Framework}
By synergistically combining the centralized allocation and decentralized execution modules, the CADE framework can be summarized by Algorithm \ref{algo:CADL}, \textcolor{black}{in which line 10 corresponds to the centralized allocation module and line 11 $\sim$ 25 correspond to the decentralized execution module in the training phase. It is worth noting that in the deployment phase when no more exploration and training is required, the decentralized execution module only includes line 11, 17, and 21.} Since we assumed that the chargers in the charging station are homogeneous, they all share the same parameter vector $\theta$. The experiences from different chargers are stored in the same replay memory. The $\epsilon$-greedy technique is deployed during the training process to ensure sufficient exploration.

\section{Numerical Study}
\label{sec:case}
In this section, we evaluate the performance of the proposed CADE framework in managing the operations of charging stations. A comprehensive comparison between the proposed and baseline algorithms in terms of optimality and scalability is conducted. We also examine the the learned action value function under different operation scenarios. The training of the proposed RL algorithm is performed on a GPU (Nvidia GeForce RTX 2080 Ti). We leveraged a desktop with AMD Ryzen 7 8-core CPU and Gurobi solver to execute the MPC-based baseline algorithm.

\subsection{Setup of the Numerical Study}
\begin{figure}[h]
\centering
\includegraphics[width=3in]{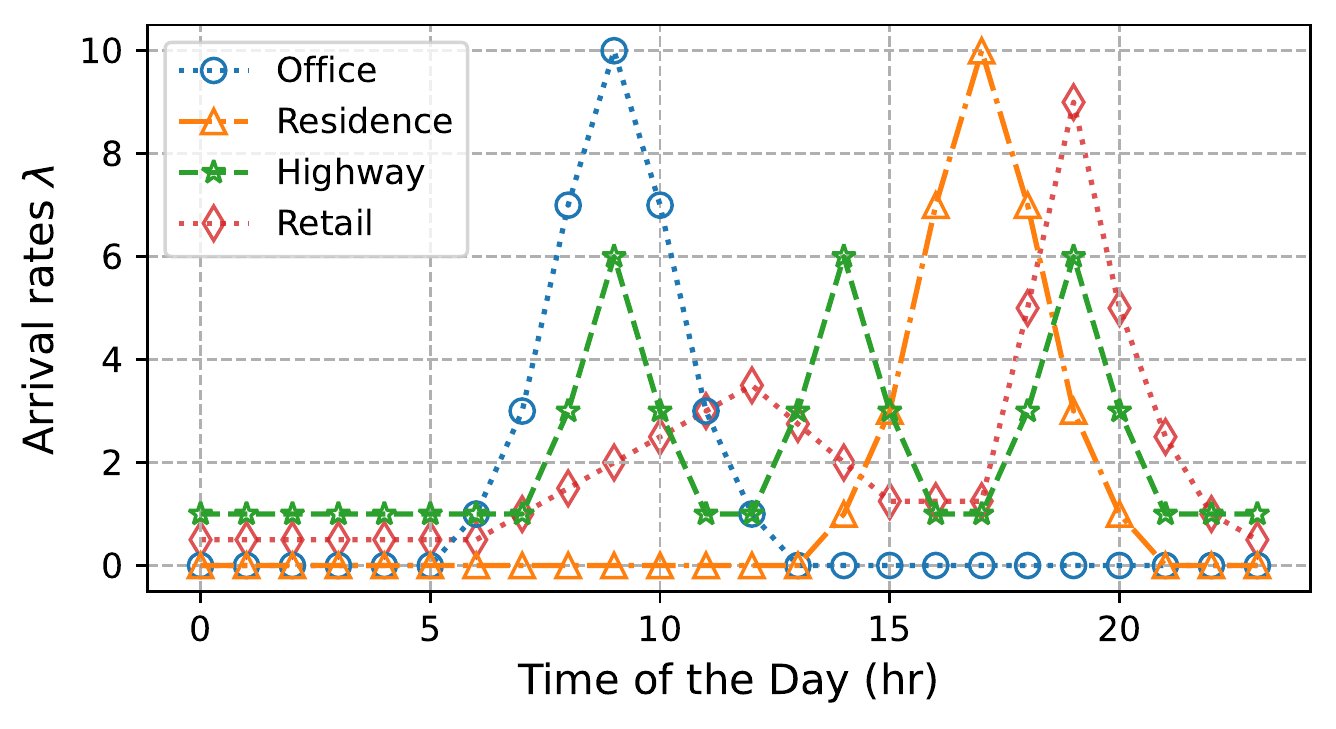}
\caption{Four different EV arrival patterns}
\label{fig:arrival}
\end{figure}

It is assumed that the arrival of EVs at the charging station follows the Poisson process. The EV arrival patterns in four different locations are evaluated: \textcolor{black}{Office zone}, residential area, highway, and \textcolor{black}{retail stores}. The arrival rates per hour $\lambda$ of the four locations are shown in Fig.~\ref{fig:arrival}. For instance, a charging station near the residential area may expect higher EV arrival rates in the afternoon. Highway charging stations may see several peaks of arrival rates in a typical day \cite{arias2017prediction} \textcolor{black}{and the retail store features a noon-peak and a higher evening peak.} \textcolor{black}{It is worth emphasizing that the proposed method does not assume prior knowledge of the specific arrival patterns.}

\begin{table}[h]
\caption{Electricity Price and Demand Charge}
\begin{center}
\begin{tabular}{c | c | c | c }
    \toprule
    %\textbf{Phase $h$} & \thead{On-peak \\ 12PM - 5PM} & \thead{Mid-peak \\ 8AM - 12PM} & \thead{Off-peak \\ 5PM - 8AM}\\
    \thead{Time-of-Use \\ Period ($h$)} & \thead{On-peak \\ 12PM - 5PM} & \thead{Mid-peak \\ 8AM - 12PM} & \thead{Off-peak \\ 5PM - 8AM}\\
    \midrule
    $p^c$ (\$/kWh) & 0.15 & 0.15 & 0.15 \\
    \midrule
    $p^d$ (\$/kWh) & 0.16 & 0.16  & 0.16\\
    \midrule
    $p^e$ (\$/kWh) & 0.20 & 0.10  & 0.05\\
    \midrule
    $p^l$ (\$/kW)  & 2.0 & 1.0  & 0.5 \\
    \bottomrule
\end{tabular}
\end{center}
\label{tab:prices}
\end{table}

\begin{table}[h]
\caption{Charging Station and Environment Parameters}
\begin{center}
\begin{threeparttable}
\begin{tabular}{c | c | c }
    \toprule
    \textbf{Parameter} & \thead{Value} & \thead{Unit} \\
    \midrule
    Number of chargers $N^c$ & 10$\sim$100 & -  \\
    \midrule
    Number of waiting spots $N^w$ & 5$\sim$50 & - \\
    \midrule
    Battery energy limits $e^{min}$/$e^{max}$ & 10/100 & kWh \\
    \midrule
    Charging power limits $a^{min}$/$a^{max}$ & -100/100 & kW \\
    \midrule
    Step change in charging power $\Delta a$ & 1 & kW \\
    \midrule
    Penalty coefficient $\mu$ & 0.2 & \$/kWh \\
    \midrule
    Time horizon $T$ & 48 & hr \\
    \midrule
    Time interval $\Delta t$ & 15 & min\\
    \midrule
    Billing period $T_B$ & 30 & day \\
    \midrule
    Dwelling time (\textcolor{black}{office}) & $\sim \mathcal{N}(8, 4^2)$ & hr \\
    \midrule
    Dwelling time (residential) & $\sim \mathcal{N}(16, 4^2)$ & hr \\
    \midrule
    Dwelling time (highway) & $\sim \mathcal{N}(1, 0.5^2)$ & hr \\
    \midrule
    \textcolor{black}{Dwelling time (retail)} & \textcolor{black}{$\sim \mathcal{N}(1, 0.75^2)$} & \textcolor{black}{hr} \\
    \midrule
    Initial battery energy level $e^{ini}$ & $\sim \mathcal{N}(20, 10^2)$ & kWh \\
    \midrule
    Target battery energy level $e^{tgt}$ & $\sim \mathcal{N}(80, 10^2)$ & kWh \\
    \bottomrule
\end{tabular}
%\begin{tablenotes}\footnotesize
%\item[*] $\sim N(\mu, \sigma)$ means the parameter is following Gaussian distribution with mean $\mu$ and standard deviation $\sigma$.
%\end{tablenotes}
\end{threeparttable}
\end{center}
\label{tab:params}
\end{table}

We assume that there are three time-of-use periods in a day: on-peak, mid-peak, and off-peak. The electricity price $p^e$ and the demand charge price $p^l$ paid by the charging station in each of the time-of-use periods are listed in Table~\ref{tab:prices}. The energy charging price $p^c$ and discharging price $p^d$ paid by EV customers are assumed to be constant. Note that we set $p^d > p^c$ to compensate the EV customers for battery degradation.

\begin{figure}[!h]
\centering
\includegraphics[width=2.9in]{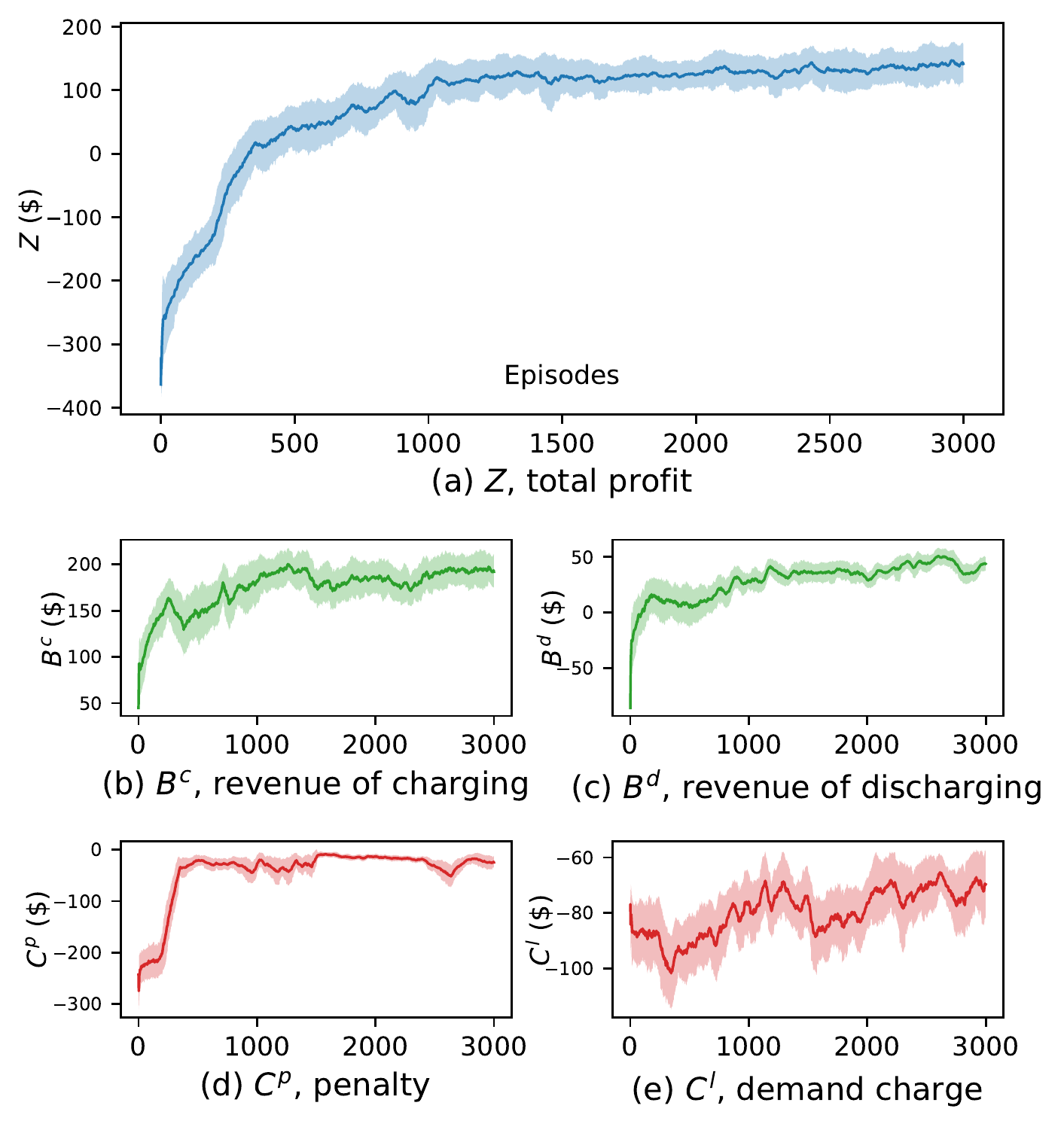}
\caption{The charging station profit breakdown in the model training process.}
\label{fig:training}
\end{figure}

\begin{figure}[!h]
\centering
\includegraphics[width=3.0in]{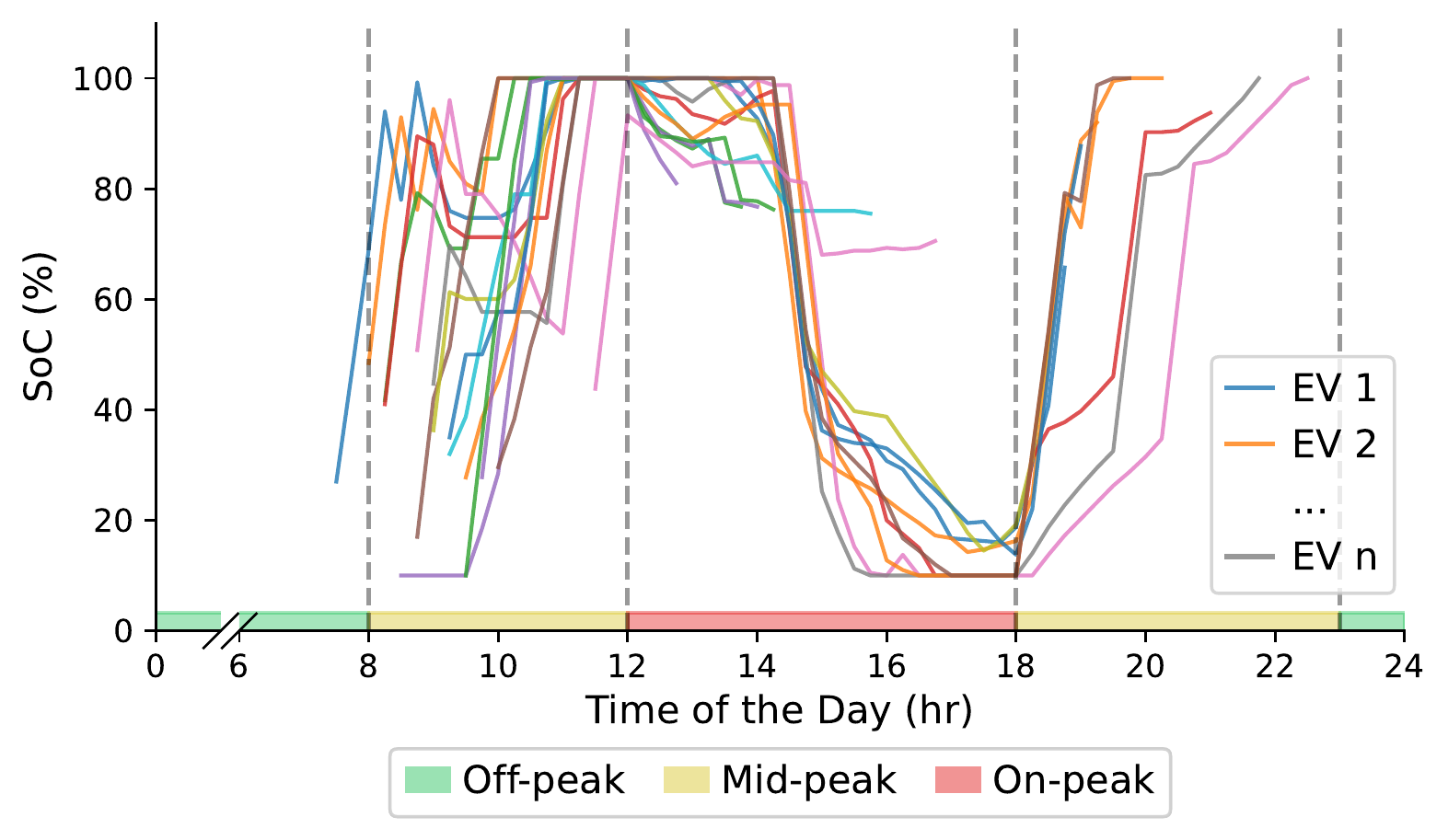}
\caption{The daily state-of-charge time series of EV batteries.}
\label{fig:energy}
\end{figure}

\begin{figure}[h]
\centering
\includegraphics[width=3.0in]{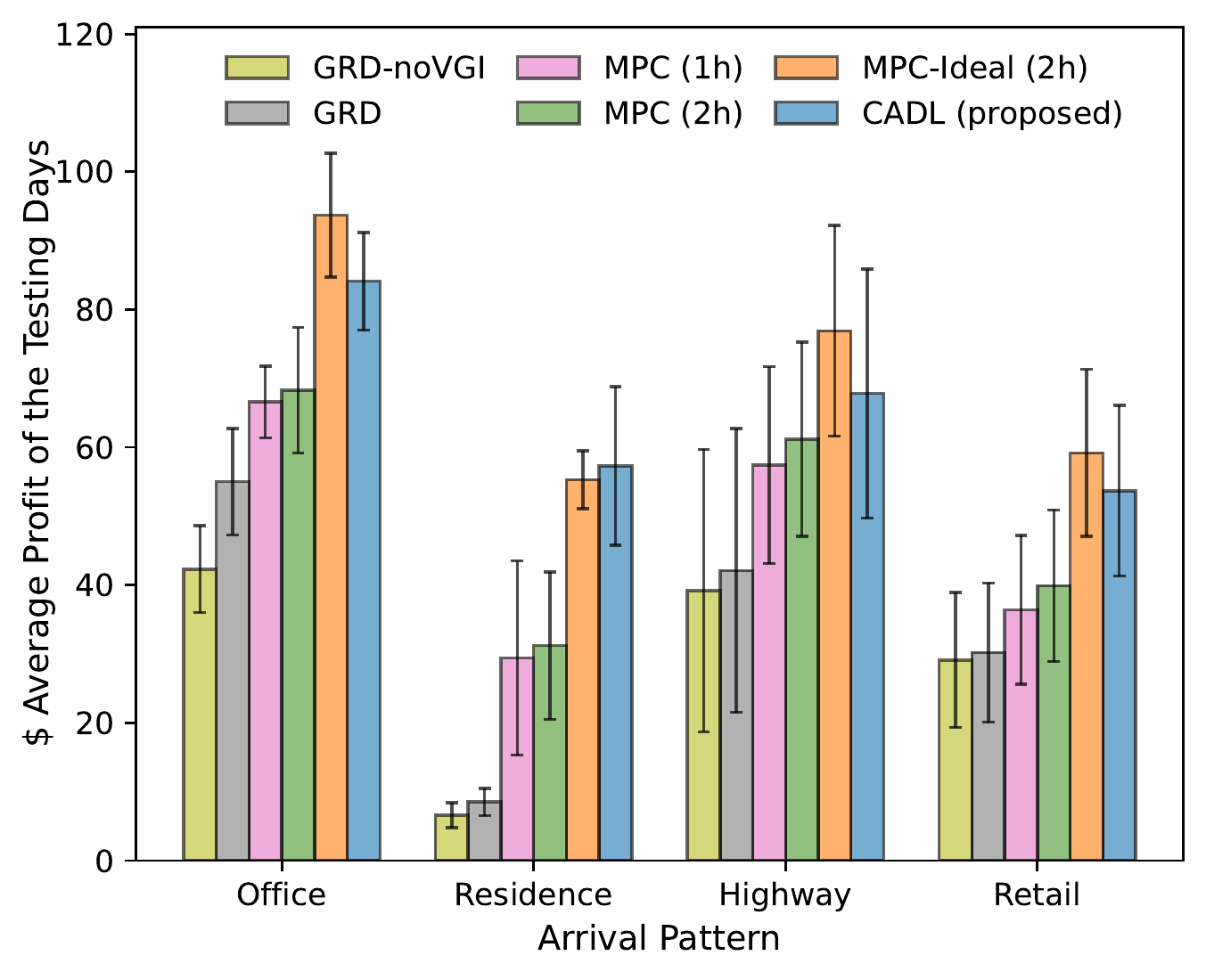}
\caption{Profit comparison between CADE \& baseline methods}
\label{fig:profit-patterns}
\end{figure}

\begin{figure}[h]
\centering
\includegraphics[width=2.8in]{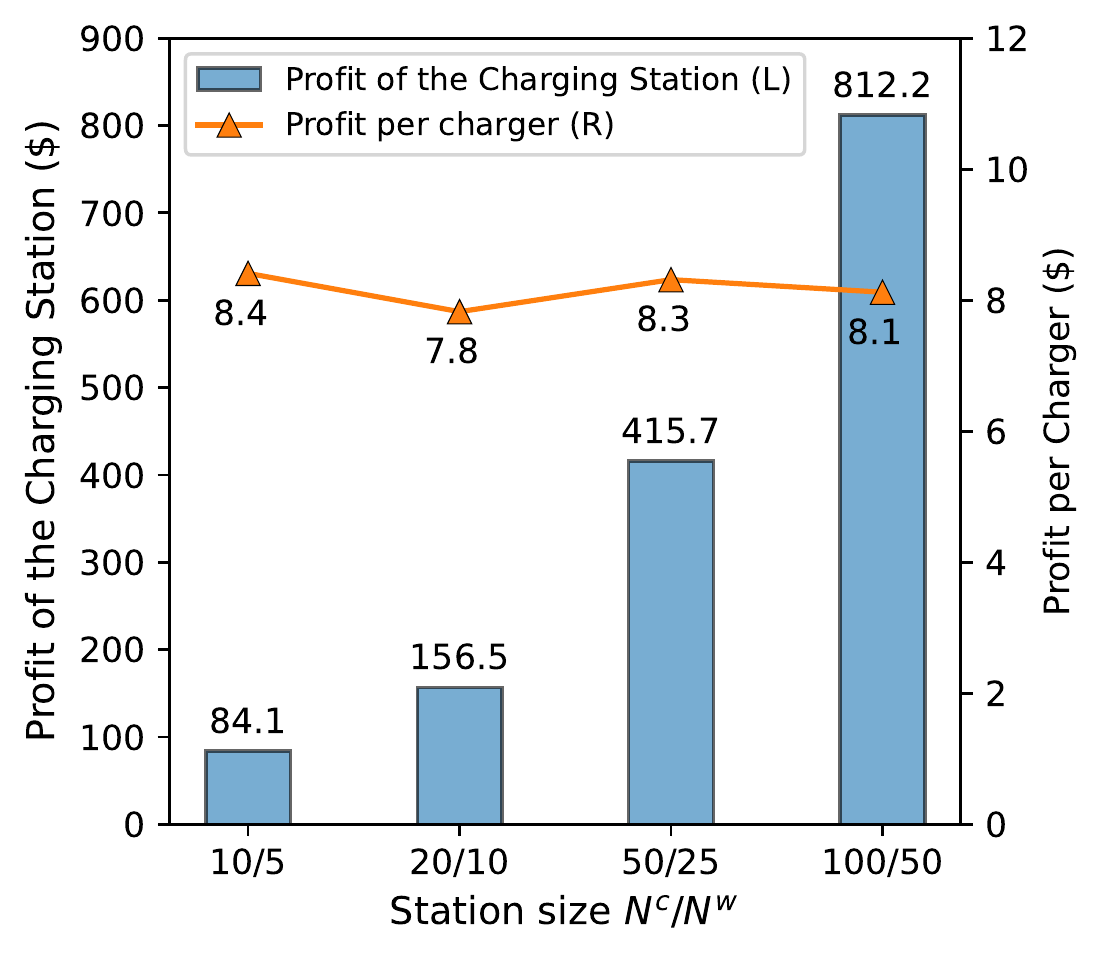}
\caption{Performance of CADE under different station sizes.}
\label{fig:profit-scale}
\end{figure}

\begin{figure*}[h]
\centering
\includegraphics[width=0.95\textwidth]{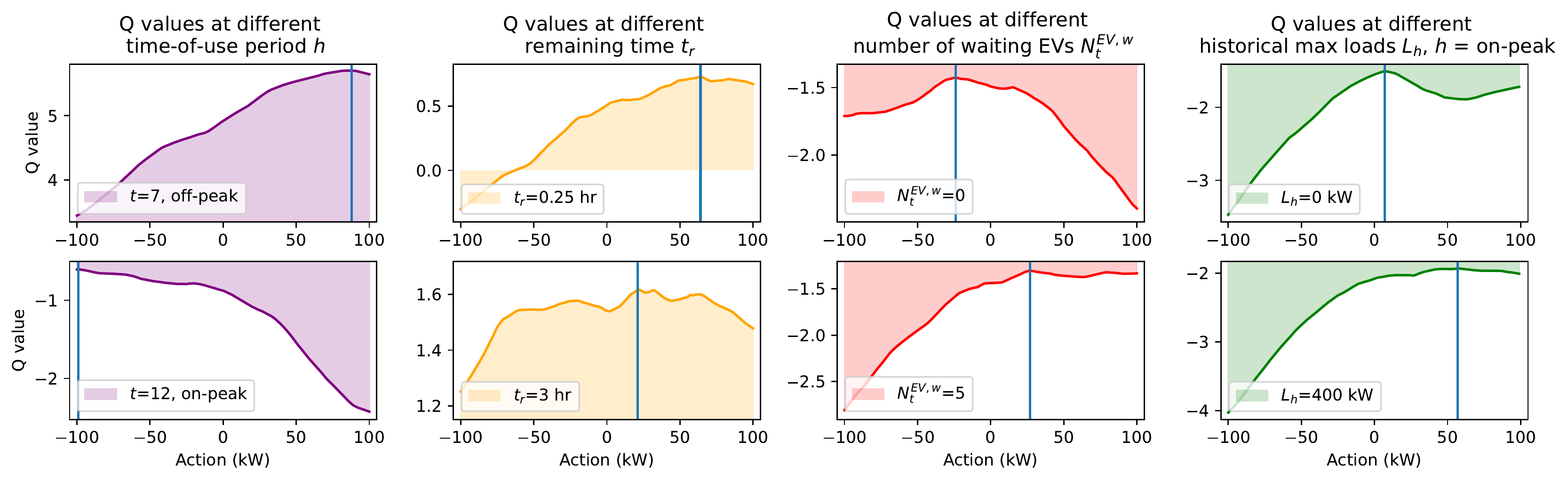}
\caption{$Q$ values of four scenarios with different (a) time-of-use period $h$, (b) remaining charging time $t_r$, (c) number of EVs in the waiting area $N^{EV,w}_t$, and (d) historical maximum loads $L_h$. }
\label{fig:q-values}
\end{figure*}

The remaining charging station and environment parameters such as size of the charging station, penalty price for not satisfying charging needs, battery energy limits, dwelling time and initial/target battery energy levels are listed in Table \ref{tab:params}.

The parameters of the RL module in the proposed CADE framework is setup as follows. The neural network approximating the action value function has three hidden layers with 256, 128, and 64 neurons, respectively. The optimizer updating Q-network parameters is stochastic gradient descent (SGD). The initial learning rate is set at 0.01 and it is scheduled to decrease along the training process. The discount factor $\gamma$ is set to be 1. The target network is updated every 25 episodes.

\subsection{Learning Performance of the Proposed CADE Framework}
The RL-module of the CADE framework is trained by following Algorithm \ref{algo:CADL}. The episodic returns as the training progresses under the office zone EV arrival pattern with $N^c=10$ and $N^w=5$ are depicted in
Fig.~\ref{fig:training}. Sub-figure (a) shows that as training progresses, the charging station profit increases steadily and then stabilizes when the number of training episodes reaches around 1,000. As shown in sub-figures (b)-(e), all components of the profit improve over the training session. The net revenues of charging ($B^c$) and discharging ($B^d$) reach high levels. The penalty term $C^p$ gradually reduces to zero, which means that the RL agents successfully learned to charge EVs to their target energy levels upon departures. The reduction of demand charge $C^l$ is limited. This is because of the difficulty to predict future EV charging demands. In sum, the CADE framework learned to improve the operation policy of the charging station.

In Fig.~\ref{fig:energy}, we visualize the daily state-of-charge time series of EVs that have visited the charging station. As shown in the figure, the chargers learned to charge EV batteries mostly during off-peak and mid-peak hours, when $p^c - p^e > 0$ and the demand charge price $p^l$ is relatively low. Similarly, the chargers learned to discharge EV batteries during on-peak hours when $p^d-p^e>0$ and $p^l$ is high.

\subsection{Profit Comparison with Baseline Methods}
We compare the charging station profit obtained by the proposed CADE framework and four baseline methods. The first baseline method is the MPC-based approach, where the one-hour \textcolor{black}{and two-hour ahead} EV arrivals are predicted based on the ground-truth Poisson distribution. The second baseline is the MPC-ideal method, which assumes the charging station has perfect prediction for \textcolor{black}{two-hour ahead} EV arrivals. \textcolor{black}{For both MPC and MPC-ideal methods, at each time step, the optimization problem (\ref{eq:op})-(\ref{eq:cons-7}) are solved by replacing the original time horizon $T$ with the prediction horizon and the solution for this time step is executed.} The third and fourth baselines are greedy methods. The GRD method assigns EVs to chargers based on urgency of charging needs. \textcolor{black}{The urgency is measured by $e^r/t^r$, the minimum charging power required to fulfill the energy demand upon an EV's departure. If an EV has a higher required minimum charging power, it has a higher urgency, which leads to a higher priority to be allocated to a charger.} The chargers then charges EVs greedily at maximal power during off-peak hours, and discharges EVs evenly during on-peak hours to a level \textcolor{black}{(no less than $E^{tgt}$)} that avoids penalty. If discharging EV battery is not allowed, then the method is denoted as GRD-noVGI. All results are obtained with $N^c=10$ and $N^w=5$. The proposed and baseline methods are evaluated using 30 randomly generated trajectories from each of the 4 different EV arrival patterns. As shown in Fig.~\ref{fig:profit-patterns}, the proposed CADE framework outperforms all baseline methods in terms of profit except MPC-ideal, which has an unfair advantage of having the perfect knowledge of future EV arrivals.

\subsection{Scalability Analysis}
In this subsection, we analyze the scalability of the proposed method. We evaluate the performance of CADE under 4 different charging station sizes with the number of chargers $N^c \in \{10, 20, 50, 100\}$. For each case, $N^w$ is kept as $0.5N^c$. The office zone EV arrival pattern is used and scaled linearly with the number of chargers. The average charging station and per charger profit under the 4 different cases are reported in Fig.~\ref{fig:profit-scale}. As the charging station size increases, the average daily profit per charger stays in a stable range from \$7.8 to \$8.4. This result shows that the proposed CADE framework can efficiently manage the operations of a large group of chargers. 

The proposed CADE framework not only outperforms baseline methods in terms of profit but also achieves high computation efficiency. The computation time of the proposed and MPC-based methods are listed in Table~\ref{tab:time}. As shown in the table, the computation time of the proposed CADE framework is much shorter than that of the MPC-based method. The decentralized execution scheme of CADE enables its computation time to increase linearly with the number of chargers rather than exponentially.

\begin{table}[h]
\caption{Comparison of computation time \textcolor{black}{per step}.}
\begin{center}
\begin{threeparttable}
\begin{tabular}{c | c | c | c | c}
    \toprule
    Station size ($N^c/N^w$) & 10/5 & 20/10 & 50/25 & 100/50\\
    \midrule
    CADE & 0.7ms & 0.8ms  & 1.1ms & 1.7ms \\
    \midrule
    MPC (1h) & 3s & 8s  & 313s  & 483s \\
    \midrule
    \textcolor{black}{MPC (2h)} & \textcolor{black}{19s} & \textcolor{black}{42s}  & \textcolor{black}{506s}  & \textcolor{black}{935s} \\
    \bottomrule
\end{tabular}
%\begin{tablenotes}\footnotesize
%\item[*] Computation time of MPC increases drastically at this and the following station sizes.
%\end{tablenotes}
\end{threeparttable}
\end{center}
\label{tab:time}
\end{table}

\subsection{Interpretation of the Learned State-Action Values}
In our proposed CADE framework, the state-action value functions are represented by neural networks, which are difficult to interpret. To better understand how the trained RL-based chargers make decisions under different scenarios, we analyze the learned state-action values. Specifically, we change one component in the state vector at a time while fixing the rest to quantify how this component impacts the distribution of $Q$ values. Fig.~\ref{fig:q-values}(a)-(d) illustrate $Q$ values of four scenarios with different (a) time-of-use-period $h$, (b) remaining charging time $t_r$, (c) number of EVs in the waiting area $N^{EV,w}_t = N^{EV}_t - N^c$, and (d) historical maximum loads $L_h$. The vertical blue lines indicate the maximum $Q$ values and the associated actions. As shown in Fig.~\ref{fig:q-values}(a), the RL-based charger prefers discharging in on-peak hours and charging in off-peak hours to increase operating profit. Fig.~\ref{fig:q-values}(b) shows that when the remaining charging time $t_r$ is close to 0, i.e. the EV is about to leave, the RL agent prefers higher charging powers in order to meet the target battery energy level. In Fig.~\ref{fig:q-values}(c), we can see that higher charging powers are preferred when there are more EVs in the waiting area. As shown in Fig.~\ref{fig:q-values}(d), if a higher historical maximum load $L_h$ is recorded, the chargers are less concerned about incurring even higher demand charge, which led to the preference of larger charging power.

\section{Conclusion}
\label{sec:conclusion}
In this paper, a centralized allocation and decentralized execution (CADE) framework is developed to operate a charging station considering vehicle-grid integration. The objective of the proposed algorithm is to maximize the profit of the charging station by considering the net revenue of charging and discharging, as well as the operational costs associated with demand charge and penalty. A centralized optimization module handles the allocation of EVs among waiting and charging spots. Enabled by the reinforcement learning algorithm, the chargers collect and share operational experiences, which are used to learn charger control policies. Comprehensive numerical studies with different EV arrival patterns show that the proposed CADE framework outperforms state-of-the-art baseline algorithms. The scalability analysis shows that the CADE framework is more computationally efficient than the baseline model-based control algorithm. Detailed analysis on the learned state-action value function provides insights on how an RL-based charger makes charging and discharging decisions under different operational scenarios.

\bibliographystyle{IEEEtran}
\bibliography{main}

\end{document}